\documentclass{article}

\usepackage{microtype}
\usepackage{graphicx}
\usepackage{booktabs} 
\usepackage{caption}
\usepackage{subcaption}

\usepackage{hyperref}

\usepackage[accepted]{icml2023}

\usepackage{amsmath}
\usepackage{amssymb}
\usepackage{mathtools}
\usepackage{amsthm}

\usepackage[capitalize,noabbrev]{cleveref}

\theoremstyle{plain}

\theoremstyle{definition}

\theoremstyle{remark}

\usepackage[textsize=tiny]{todonotes}

\usepackage{amsmath,amsfonts,bm}

\newcommand{\defined}{\overset{\operatorname{def}}{=}}

\def\eqref#1{equation~\ref{#1}}

\def\1{\bm{1}}

\def\rx{{\textnormal{x}}}

\def\rvx{{\mathbf{x}}}
\def\rvy{{\mathbf{y}}}

\def\vx{{\bm{x}}}
\def\vy{{\bm{y}}}

\DeclareMathAlphabet{\mathsfit}{\encodingdefault}{\sfdefault}{m}{sl}
\SetMathAlphabet{\mathsfit}{bold}{\encodingdefault}{\sfdefault}{bx}{n}

\def\gF{{\mathcal{F}}}

\def\gQ{{\mathcal{Q}}}
\def\gR{{\mathcal{R}}}

\def\sR{{\mathbb{R}}}

\def\sX{{\mathbb{X}}}
\def\sY{{\mathbb{Y}}}

\newcommand{\E}{\mathbb{E}}

\newcommand{\KL}{{\mathrm{KL}}}
\newcommand{\CHI}{{\chi^2}}
\newcommand{\Var}{\mathbb{V}}

\newcommand{\lunnorm}[3]{{\KL_{#1}(#2||#3)}}

\DeclareMathOperator*{\argmax}{arg\,max}

\usepackage{soul,xcolor}

\icmltitlerunning{On the Effectiveness of Hybrid Mutual Information Estimation}

\begin{document}

\twocolumn[
\icmltitle{On the Effectiveness of Hybrid Mutual Information Estimation}




\begin{icmlauthorlist}
\icmlauthor{Marco Federici}{uva}
\icmlauthor{David Ruhe}{uva}
\icmlauthor{Patrick Forr\'e}{uva}
\end{icmlauthorlist}

\icmlaffiliation{uva}{AMLab, University of Amsterdam}

\icmlcorrespondingauthor{Marco Federici}{m.federici@uva.nl}

\icmlkeywords{Machine Learning, Information Theory}

\vskip 0.3in
]
\definecolor{ggreen}{RGB}{58, 146, 58}


\printAffiliationsAndNotice{} 

\begin{abstract}
    \begin{enumerate}
    \end{enumerate}
        Estimating the mutual information from samples from a joint distribution is a challenging problem in both science and engineering.
        In this work, we realize a variational bound
         that generalizes both discriminative and generative approaches.
         Using this bound, we propose a hybrid method 
         to mitigate their respective shortcomings.
        Further, we propose Predictive Quantization (PQ): a simple generative method that can be easily combined with discriminative estimators for minimal computational overhead.
        Our propositions yield a tighter bound on the information thanks to the reduced variance of the estimator.
        We test our methods on a 
        challenging task of correlated high-dimensional Gaussian distributions and a stochastic process involving a system of free particles subjected to a fixed energy landscape.
        Empirical results show that hybrid methods consistently improved mutual information estimates when compared to the corresponding discriminative counterpart. 
\end{abstract}

\section{Introduction}
   \begin{figure*}
    \centering
    \begin{minipage}{0.19\textwidth}
    \centering
    $\log p(\rvx)p(\rvy)$
    \includegraphics[width=\textwidth]{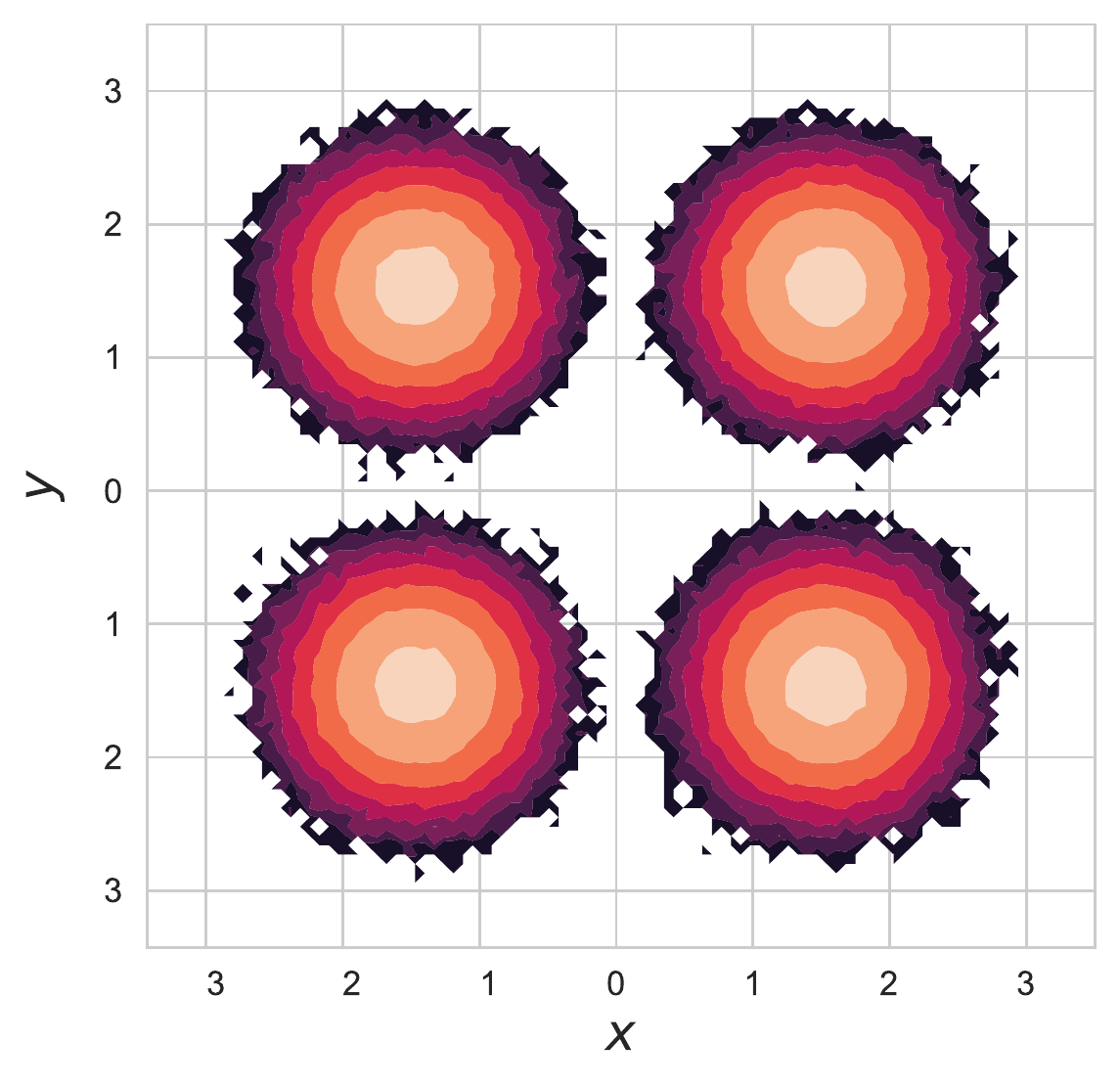}
    \end{minipage}
    \begin{minipage}{0.19\textwidth}
    \centering
    $\log p(\rvx)p(\rvy|Q_2(\rvx))$
    \includegraphics[width=\textwidth]{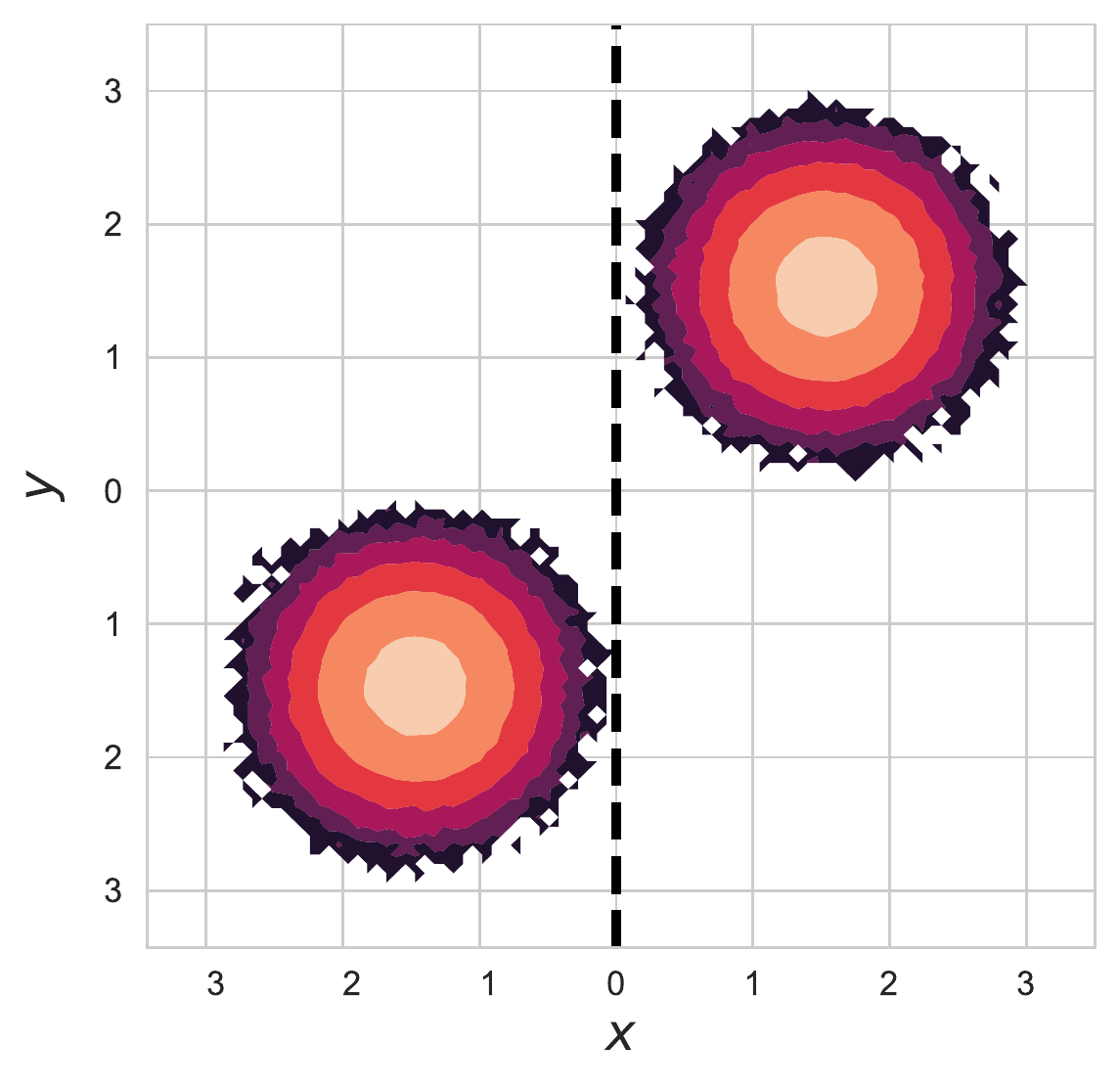}
    \end{minipage}
    \begin{minipage}{0.19\textwidth}
    \centering
    $\log p(\rvx)p(\rvy|Q_4(\rvx))$
    \includegraphics[width=\textwidth]{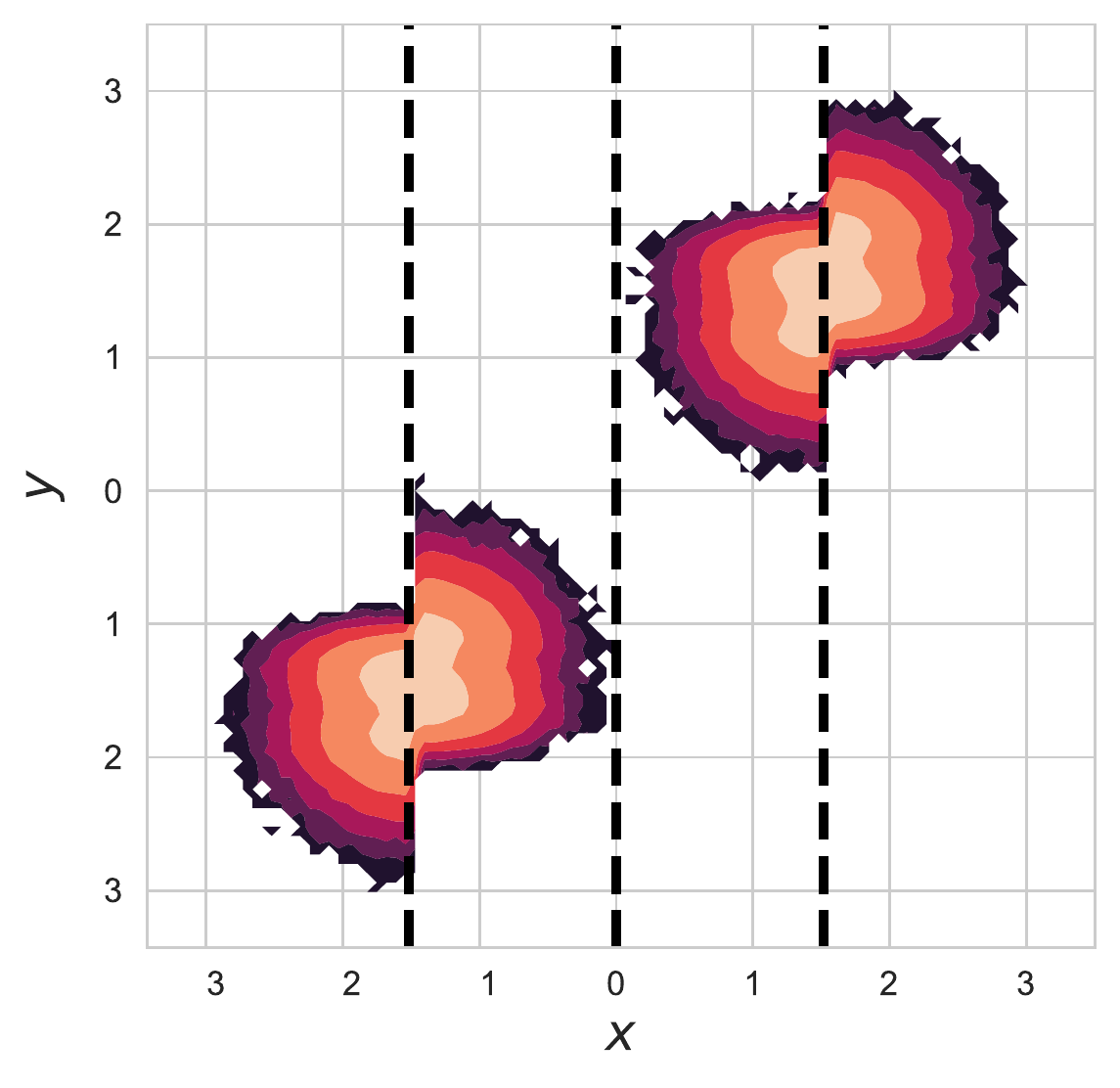}
    \end{minipage}
    \begin{minipage}{0.19\textwidth}
    \centering
    $\log p(\rvx)p(\rvy|Q_8(\rvx))$
    \includegraphics[width=\textwidth]{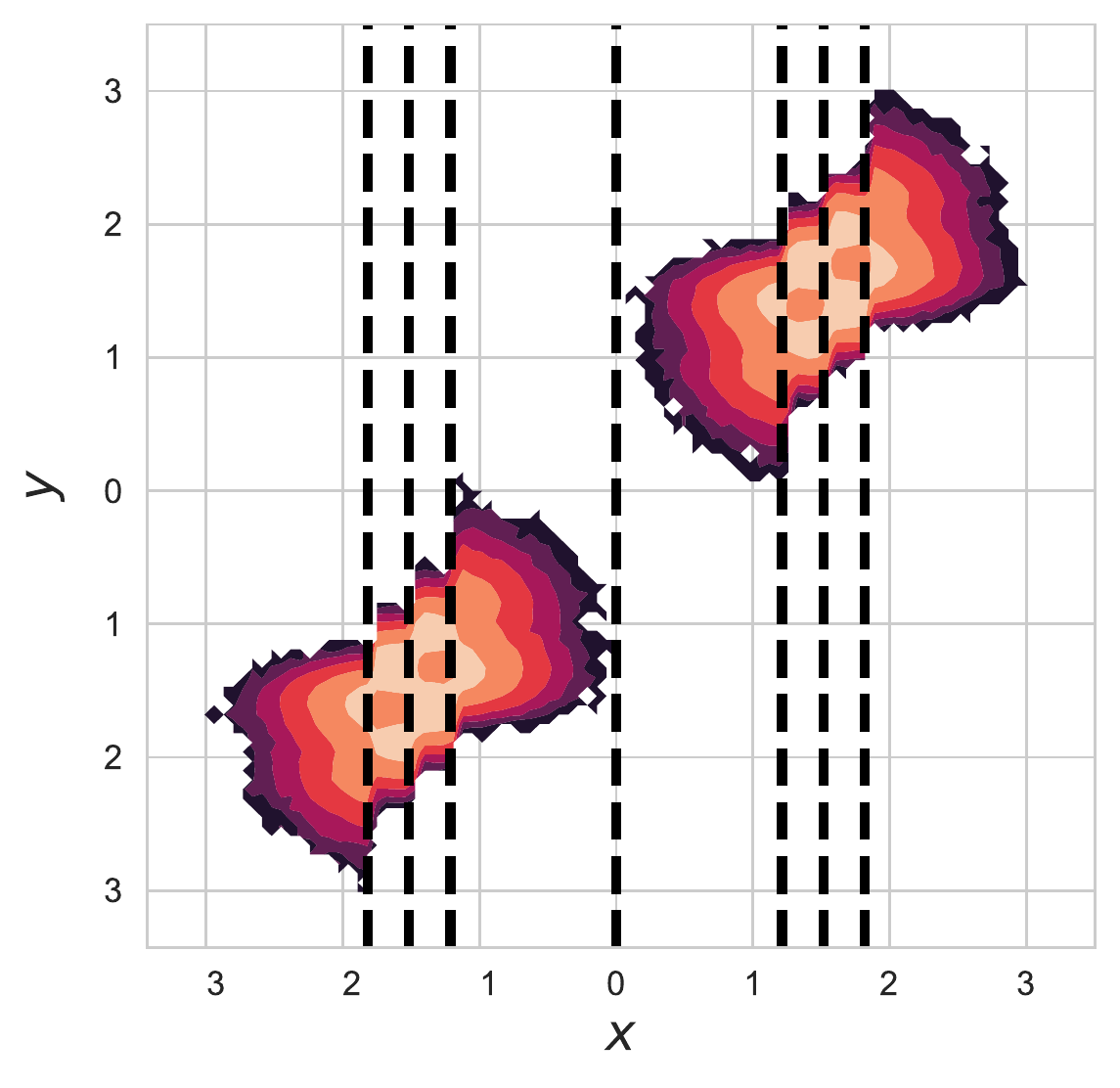}
    \end{minipage}
    \begin{minipage}{0.19\textwidth}
    \centering
    $\log p(\rvx)p(\rvy|\rvx)$
    \includegraphics[width=\textwidth]{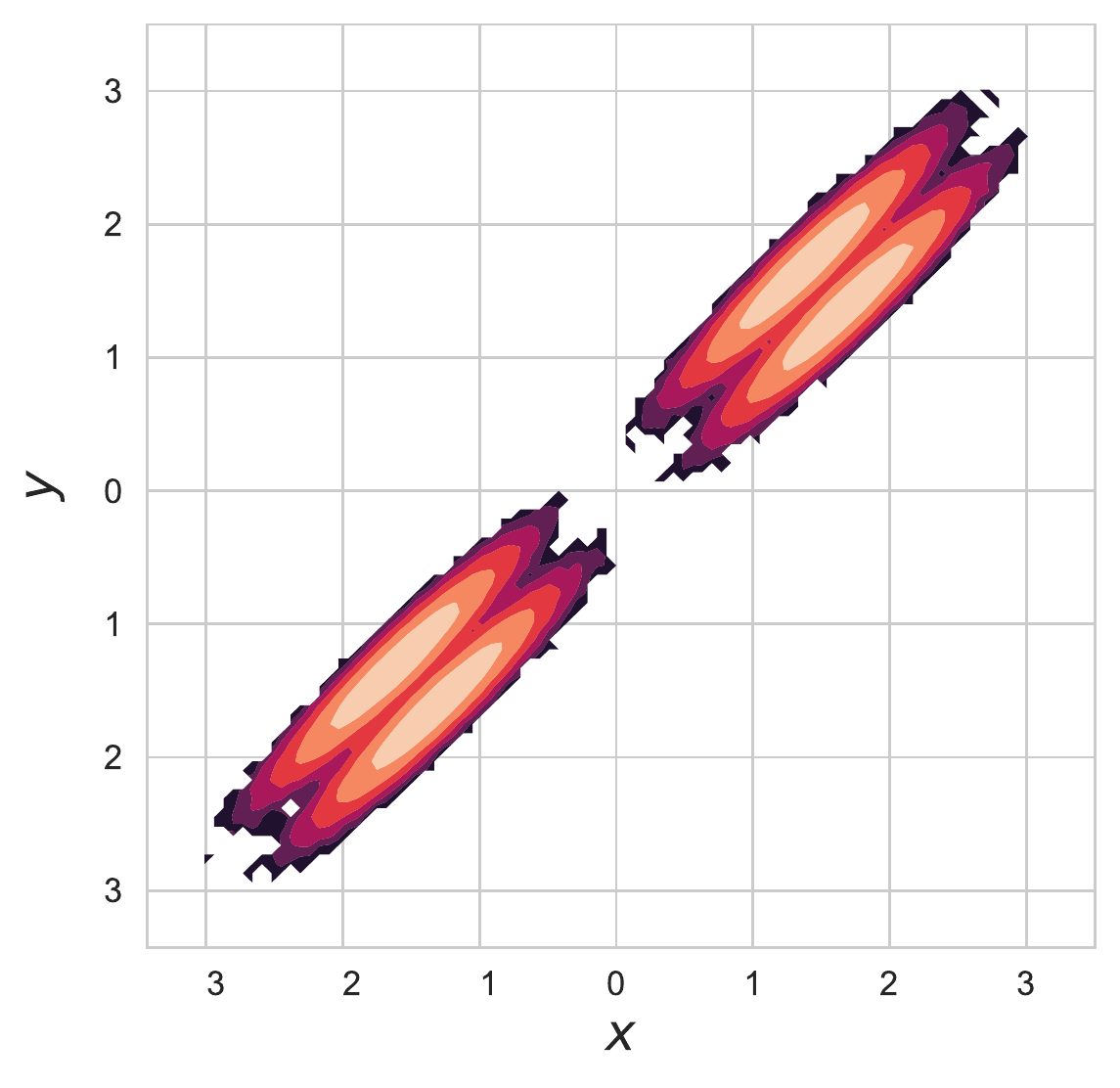}
    \end{minipage}
    \caption{Predictive Quantization (PQ). 
    This figure shows how a simple quantization function $Q(\rvx)$ can be used to obtain proposals that lie closer to a target joint distribution (on the right) than the product of the marginals (on the left).
    We can use these intermediate distributions to split the problem of estimating mutual information into two parts, effectively reducing the variance of popular discriminative estimators.
    This results in a tighter estimate of the variational bound.
    Vertical dashed lines are used to indicate the quantized regions. Note that as the number of partitions increases, the proposal approaches the joint distribution.
    }
    \label{fig:attribute_effect}
\end{figure*}
Mutual Information quantifies the amount of information 
gained about one random variable by observing another one \citep{shannon1948mathematical, mackay2003information}.
As such, it is a measure of their mutual dependence.
Estimating the dependency of two random variables is a problem found ubiquitously in science and engineering.
Examples are independent component analysis \citep{hyvarinen2000independent, bach2002kernel}, neuroscience \citep{palmer2015predictive}, Bayesian experimental design \citep{ryan2016review}.
Further, mutual information estimation plays a crucial role in self-supervised learning, which in recent years has gained significant attention from the machine learning community \citep{hjelm2018learning, zbontar2021barlow} in the form of information optimization.
Information optimization, such as in the information bottleneck method (minimization) \citep{tishby2000information, alemi2016deep} or deep representation learning \citep{bengio2013representation, hjelm2018learning, oord2018representation}, can be conducted without explicitly modeling the underlying distributions.
Instead, through directly bounding the mutual information, one can obtain tractable objective functions that can be optimized using flexible function estimators.

Mutual information estimation, the core focus of this work, is a challenging task since one usually has access only to samples from the underlying probability distributions, and not their densities \citep{paninski2003estimation, mcallester2020formal}. 
Classical estimators \citep{kraskov2004estimating, gao2015efficient} typically break down when the data is higher-dimensional and can be non-differentiable, making them unsuitable for gradient-based information optimization \citep{hjelm2018learning}.

Variational mutual information estimation considers 
a lower bound on the true information.
Therefore, maximizing this bound using a flexible variational family yields accurate information estimates.
By identifying a bound that unifies the generative and discriminative methods as categorized by \citet{song2020understanding} and \citet{poole2019variational}, we aim to address 
their respective shortcomings.
That is, generative approaches lack flexibility and discriminative approaches suffer from unfavorable bias-variance trade-offs for large estimates.
The unification of these techniques provides a method that can provide more accurate estimation.

Further, we specify a simple yet effective generative method called \emph{Predictive Quantization} (PQ) that makes use of a quantized (discrete) representation of the data, such as a clustering or a set of external attributes, to approximate quantities (such as a marginal entropy) that are usually intractable.

Our contributions, therefore, include the following: 
\begin{enumerate}
    \item We introduce a novel family of hybrid mutual information estimators which generalizes both generative and discriminative approaches.
    \item We show theoretically and empirically that there is a clear advantage to combining generative and discriminative mutual information estimators since they have complementary strengths and weaknesses.
    \item We design Predictive Quantization (PQ): a simple generative method that can to improve a wide range of recently proposed discriminative estimators. 
\end{enumerate}

Experimentally, we test our models by estimating mutual information between the dimensions of a mixture of correlated multi-dimensional Gaussian distributions and compare the estimates of temporal information for a stochastic process of spatially correlated moving particles subjected to a fixed energy landscape.

\section{Mutual Information Estimation}

Given two random variables $\rvx$ and $\rvy$ with support $\sX$, $\sY$ and joint probability density $p(\rvx, \rvy)$, the mutual information between $\rvx$ and $\rvy$ is defined as Kullback-Leibler (KL) divergence between $p(\rvx, \rvy)$ and the product of the two marginal distributions\footnote{Unless otherwise specified, expectations are computed with respect to $p(\rvx,\rvy)$. See \Cref{app:notation} for notational details.}:
\begin{align}
    I(\rvx;\rvy)&\defined \KL(p(\rvx,\rvy)||p(\rvx)p(\rvy))\nonumber\\
    &=\mathbb{E}\left[\log  \frac{p(\vx,\vy)}{p(\vx)p(\vy)}\right].
    \label{eq:mi_def}
\end{align}
In most applications of interest, we can sample $\vx, \vy \sim p(\rvx, \rvy)$ to compute a Monte Carlo estimate.
However, some of the densities in \cref{eq:mi_def} are usually unknown or intractable.
For this reason, a common strategy involves the introduction of a variational distribution  $q(\rvx,\rvy)$ to obtain a lower bound on mutual information, which can be used for either estimation or maximization:
\begin{align}
    I(\rvx;\rvy)&=\mathbb{E}\left[\log\left( \frac{q(\vx,\vy)}{p(\vx)p(\vy)}\frac{p(\vx,\vy)}{q(\vx,\vy)}\right)\right]\nonumber\\
    &= \mathbb{E}\left[\log \frac{q(\vx,\vy)}{p(\vx)p(\vy)}\right]+\underbrace{\KL(p(\rvx,\rvy)||q(\rvx,\rvy))}_{\text{Variational Gap}}\nonumber\\
    &\ge \mathbb{E}\left[\log \frac{q(\vx,\vy)}{p(\vx)p(\vy)}\right]\nonumber\\
    &\defined I_{q}(\rvx,\rvy),
    \label{eq:mi_lbound}
\end{align}
where $q(\rvx,\rvy)$ is a joint density in a set $\gQ$ of attainable variational densities with positive support on $\sX\times\sY$.
Note that the bound in \Cref{eq:mi_lbound} is tight only when $p(\rvx,\rvy)\in \mathcal{Q}$. 
Therefore, accurate mutual information estimation requires access to a flexible variational family. 
The KL divergence between the joint distribution $p(\rvx,\rvy)$ and its variational approximation $q(\rvx,\rvy)$ is commonly referred to as the \emph{variational gap}, which we seek to minimize for accurate estimation.

Previous work \citep{poole2019variational, song2020understanding} has identified two categories of approaches for maximizing the ratio $I_q(\rvx;\rvy)$: generative and discriminative methods.
In the following sections, we recover both these approaches using a parameterization for $q(\rvx, \rvy)$ consisting of a normalized proposal distribution (generative) and an unnormalized density ratio (discriminative).
On one hand, discriminative approaches \citep{belghazi2018mutual, nguyen2010estimating} directly model the entire ratio, but this comes at the cost of 
considerable bias \citep{oord2018representation} or high variance for large values of mutual information \citep{mcallester2020formal}.
Generative approaches \citep{barber2003im}, on the other hand, model the components in \Cref{eq:mi_lbound} using learnable normalized densities.
However, access to flexible parameterized distributions can be hard to attain, or they are expensive to optimize. These issues usually become more severe for high-dimensional $\rvx$ and $\rvy$.
\begin{table}[]
    \centering
    \small{
    \begin{tabular}{c|c|c}
         &  Proposal & Energy \\\hline
        Generative & $r_\theta(\rvx,\rvy)$ & $f_k(\rvx,\rvy) = k$ \\
        Discriminative & $p(\rvx)p(\rvy)$ & $f_\phi(\rvx,\rvy)$ \\
        Hybrid (Ours) & $r_\theta(\rvx,\rvy)$ & $f_\phi(\rvx,\rvy)$
    \end{tabular}
    }
    \caption{Summary of the modeling choices for the main families of approaches for mutual information estimation with respect to the expression in \Cref{eq:general_bound}. Our proposed hybrid approach can exploit the flexibility of discriminative models while mitigating the variance because of the use of more expressive proposals.}
    \label{tab:approaches}
\end{table}

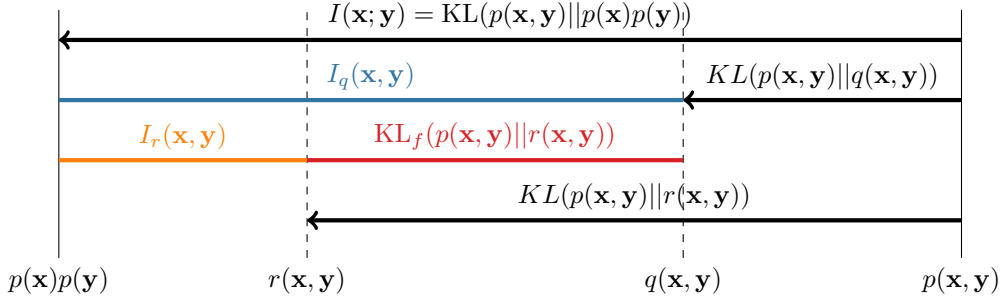
\begin{figure*}
    \centering
    \definecolor{bblue}{RGB}{50,116,161}
\definecolor{oorange}{RGB}{255, 128, 8}
\definecolor{rred}{RGB}{216, 33, 36}

\begin{tikzpicture}
\def\dy{0.8}

\def\px{12}
\def\qx{8.3}
\def\rx{3.3}
\def\ppx{0}
\node[] at (\px,0) (px)  {$p(\rvx,\rvy)$};
\node[] at (\qx,0) (qx)  {$q(\rvx,\rvy)$};
\node[] at (\rx,0) (rx)  {$r(\rvx,\rvy)$};
\node[] at (\ppx,0) (ppx)  {$p(\rvx)p(\rvy)$};

\draw[] (px) -- (\px,4.5*\dy);
\draw[dashed] (qx) -- (\qx,4.5*\dy);
\draw[dashed] (rx) -- (\rx,4.5*\dy);
\draw[] (ppx) -- (\ppx,4.5*\dy);

\draw[ultra thick, -to] (\px,4*\dy) --node[midway,above]{$I(\rvx;\rvy)=\KL(p(\rvx,\rvy)||p(\rvx)p(\rvy))$} (\ppx,4*\dy);

\draw[ultra thick, black, -to] (\px,3*\dy) -- node[midway,above]{$KL(p(\rvx,\rvy)||q(\rvx,\rvy))$} (\qx,3*\dy);
\draw[ultra thick, bblue] (\ppx,3*\dy) --node[midway,above]{$I_{q}(\rvx,\rvy)$} (\qx,3*\dy);

\draw[ultra thick, oorange] (\ppx,2*\dy) --node[midway,above]{$I_{r}(\rvx,\rvy)$} (\rx,2*\dy);
\draw[ultra thick,rred] (\qx,2*\dy) --node[midway,above]{$\lunnorm{f}{p(\rvx,\rvy)}{r(\rvx,\rvy)}$} (\rx,2*\dy);

\draw[ultra thick, black, -to] (\px,1*\dy) --node[midway,above]{$KL(p(\rvx,\rvy)||r(\rvx,\rvy))$} (\rx,1*\dy);

\end{tikzpicture}
    \caption{Visualization of the additive decomposition of the terms of \Cref{eq:mi_lbound} and \Cref{eq:general_bound}.
    The mutual information lower-bound for the implicitly defined distribution $q(\rvx,\rvy)$ (in \textcolor{bblue}{blue}) can be seen as the sum of the lower-bound for the proposal distribution $r(\rvx,\rvy)$ (in \textcolor{oorange}{orange}) and the estimation of the corresponding variational gap (in \textcolor{rred}{red}).}
    \label{fig:loss}
\end{figure*}

\section{A Generalized Variational Approach}

In the following, we set up a parametrization for $q(\rvx, \rvy)$ that allows us to derive popular mutual information estimators as special cases and propose new generalized hybrid estimators. 
First, note that any joint probability distribution $q(\rvx,\rvy)$ can be written as a normalized exponential:
\begin{align}
    q(\rvx,\rvy) &= \frac{e^{F(\rvx,\rvy)}}{Z_F},\ \ 
    \text{with } Z_F\defined \iint e^{F(\vx,\vy)} d\vx d\vy.
\end{align}
The function $F: \sX\times\sY\to\sR$ is commonly referred to as the negative \textit{energy function} and $Z_F$ as its normalization constant\footnote{The energy needs to satisfy mild integrability constraints.}.
The main advantage is that there are few restrictions on $F$, a property that has fruitfully been exploited by energy-based models \cite{song2021train, grathwohl2019your, haarnoja2017reinforcement, song2020sliced, hyvarinen2005estimation, ngiam2011learning}.

The main disadvantage, however, is that the normalization constant $Z_F$ is generally intractable due to computationally expensive integration over the support $\sX\times\sY$.
To address this problem, we compute a Monte Carlo estimate of $Z_F$ by sampling from a proposal distribution $r(\rvx, \rvy)$ from an attainable family of normalized densities $\gR$:
\begin{align}
    Z_F &= \iint\frac{r(\vx,\vy)}{r(\vx,\vy)} e^{F(\vx, \vy)} d\vx d\vy\nonumber\\
    &=\mathbb{E}_{r}\left[e^{F(\vx, \vy)-\log r(\vx,\vy)}\right].
\end{align}
Lastly, we can reparameterize the energy such that $F(\rvx,\rvy)\defined\log r(\rvx, \rvy) + f(\rvx, \rvy)$ to obtain
\begin{align}
    q(\rvx,\rvy)=r(\rvx,\rvy)\frac{e^{f(\rvx,\rvy)}}{Z_f}, \ \ \text{with }Z_f=\E_{r}[e^{f(\vx,\vy)}].
\end{align}
Intuitively, $q(\rvx,\rvy)$ is obtained by transforming the proposal $r(\rvx,\rvy)\in\gR$ with a \textit{critic} function $f\in\gF$, and re-normalizing to obtain a valid density. 
We denote the family of variational distributions obtained with this procedure as $\gQ\defined\gR_\gF$ to underline the dependency on the chosen proposals $\gR$ and critics $\gF$.

Using this parameterization in \cref{eq:mi_lbound}, we obtain a general bound that includes both a generative and a discriminative component:
\begin{align}
    I_{q}(\rvx,\rvy)&=\underbrace{\E\left[\log\frac{r(\vx,\vy)}{p(\vx)p(\vy)}\right]}_{I_{r}(\rvx,\rvy)} \nonumber\\
    &\ \ \ + \underbrace{\E[f(\vx,\vy)] - \log\E_{r}\left[e^{f(\vx,\vy)}\right]}_{\lunnorm{ f}{p(\rvx,\rvy)}{r(\rvx,\rvy)}}.
     \label{eq:general_bound}
\end{align}
The entire decomposition is visualized in \Cref{fig:loss}.
The total information $I(\rvx; \rvy)$ (first row) decomposes through \Cref{eq:mi_lbound} into the lower-bound $I_q(\rvx;\rvy)$ and respective variational gap $\KL(p(\rvx, \rvy)||r(\rvx, \rvy))$ (second row).
$I_q(\rvx;\rvy)$ is then further split into the terms of \Cref{eq:general_bound} (third row).
The first component $ I_r(\rvx;\rvy)$ is a mutual information lower bound determined by the proposal $r(\rvx,\rvy)\in\gR$. 
This quantity differs from the target mutual information $I(\rvx;\rvy)$ by the variational gap for $r(\rvx,\rvy)$: $\KL(p(\rvx,\rvy)||r(\rvx,\rvy))$ (fourth row).
I.e., $I(\rvx; \rvy) = I_r(\rvx; \rvy) + \KL(p(\rvx,\rvy)||r(\rvx,\rvy))$.

The second component in \Cref{eq:general_bound} consists of the comparison between the value of the critic $f$ of samples from the true joint to samples from the proposal.
This expression can be seen as the Donsker-Varadhan representation of the aforementioned variational gap of the proposal \citep{donsker1983asymptotic}:
\begin{align}
      \KL(p(\rvx,\rvy)||r(\rvx,\rvy))\ge\lunnorm{f}{p(\rvx,\rvy)}{r(\rvx,\rvy)},
\end{align}
in which the inequality is tight when $\gF$ contains $f^*_k(\rvx,\rvy)\defined\log\frac{p(\rvx,\rvy)}{r(\rvx,\rvy)}+k$ with $k\in\sR$ \citep{poole2019variational}.
Note that the variational gap for $q$ is always smaller than the variational gap for $r$ whenever $\gF$ contains at least a constant ($k$) function:
\begin{align}
    \exists f \in \mathcal{F},\,f: \sX \times \sY \to \{k\}  \implies \max_{f\in\gF} I_q(\rvx;\rvy) \ge I_r(\rvx;\rvy)
\end{align}
Since critics are usually modeled using flexible function approximators, this is a reasonable assumption, and, in practice, the family of transformed densities $\gR_\gF$ is much larger than original proposals $\gR$.

As summarized in \Cref{tab:approaches}, previous approaches for mutual information estimation can be seen as instances of the parameterization reported in \Cref{eq:mi_lbound} obtained by either restricting $r(\rvx, \rvy) = p(\rvx)p(\rvy)$ (discriminative methods)
or using a constant critic function (generative methods). 
In the following sections, we analyze the strengths and weaknesses of each method to design a novel hybrid approach that takes advantage of the best characteristics of both.

\subsection{The Discriminative Approach: a Bias-Variance Trade-Off}
In this section we focus on discriminative methods, which we recover by setting $r(\rvx, \rvy) = p(\rvx)p(\rvy)$, causing $I_r(\rvx, \rvy)$ to vanish from \Cref{eq:general_bound}.
On the one hand, modeling density ratios directly with neural network critics allow for great flexibility \citep{gutmann2010noise, oord2018representation}. 
On the other, the efficacy of techniques to estimate $\lunnorm{ f}{p(\rvx,\rvy)}{r(\rvx,\rvy)}$ heavily depends on bias-variance trade-offs of the chosen approximation to evaluate the normalization constant \citep{poole2019variational, mcallester2020formal, song2020understanding}. 
In practice, Monte Carlo estimation of the log-normalization constant $\log Z_f$ with a limited number of samples yields biased results caused by the log-expectation \citep{belghazi2018mutual, oord2018representation}.
For this reason, we consider a looser lower-bound of $\KL(p(\rvx,\rvy)||r(\rvx,\rvy))$ corresponding to the dual representation of the KL-divergence \citep{donsker1983asymptotic}:
\begin{align}
   &\lunnorm{f}{p(\rvx,\rvy)}{r(\rvx,\rvy)}\nonumber\\
   &\ \ \ \ge\E_{p}[f(\vx,\vy)] - \E_{r}\left[e^{f(\vx,\vy)}\right]+1,
   \label{eq:kl_dual}
\end{align}
where the inequality is tight when the normalization constant $Z_f=1$.

Estimating \Cref{eq:kl_dual} using Monte Carlo samples suffers from less bias at the cost of a higher variance \citep{nguyen2010estimating, poole2019variational, guo2022tight}.
In fact, the variance of $\lunnorm{f}{p(\rvx,\rvy)}{r(\rvx,\rvy)}$ for a critic $f\in\gF$ is mostly determined by the variance of $e^{f(\vx, \rvy)}$, which can be bounded from below by an exponential of the KL-divergence between the implicitly defined $q(\rvx,\rvy)$ and original proposal $r(\rvx,\rvy)$: 
\begin{align}
    \Var_{r}\left[e^{f(\vx,\vy)}\right]&\ge Z_f^2\CHI(q(\rvx,\rvy)||r(\rvx,\rvy))\nonumber\\
    &\ge Z^2_f\left(e^{\KL(q(\rvx,\rvy)||r(\rvx,\rvy))}-1\right).
    \label{eq:variance}
\end{align}
Here, $\CHI(q(\rvx,\rvy)||r(\rvx,\rvy))$ denotes the Pearson chi-squared divergence between the variational distribution and the proposal.
Intuitively, the variance increases the more the critic changes the proposal.
Whenever the critic is constant on the whole support $\sX\times\sY$, $q(\rvx,\rvy)$ matches $r(\rvx,\rvy)$, rendering the variance and value of $\lunnorm{f}{q(\rvx,\rvy)}{r(\rvx,\rvy)}$ zero.
In this setting, the variational gap is completely determined by the proposal, since $I_q(\rvx;\rvy)=I_r(\rvx;\rvy)$. 
In contrast, when $q(\rvx,\rvy)=p(\rvx,\rvy)$ 
(i.e., the critic is optimal $f(\rvx, \rvy) = f^*(\rvx,\rvy)\defined\log\frac{p(\rvx,\rvy)}{r(\rvx,\rvy)}$), the variational gap for $q(\rvx,\rvy)$ is zero, but the variance is still determined by an exponential of the variational gap for $r(\rvx,\rvy)$.
In the case in which $r(\rvx,\rvy) = p(\rvx)p(\rvy)$, the variance grows exponentially with the amount of information to estimate \citep{mcallester2020formal, song2020understanding}.

\subsection{On the Generative Approach}
When setting $f(\rvx, \rvy)=k$, the second term of \Cref{eq:general_bound} vanishes.
We are now in the generative setting. 
Considering $I_r(\rvx; \rvy)$, computing and optimizing the ratio between the proposal and the product of the marginals requires access to a flexible family of distributions with known (or approximate) densities. 
Depending on the modeling choices for $r(\rvx,\rvy)$, the computation may require estimating up to two entropy and one cross-entropy term:
\begin{align}
    I_r(\rvx;\rvy) &= \E[\log r(\rvx,\rvy)] + H(\rvx) + H(\rvy).
\end{align}
\citet{barber2003im} and \citet{mcallester2020formal} model a joint proposal as the product of one of the marginals  and a conditional proposal $r_\rvx(\rvx,\rvy)\defined p(\rvx)r(\rvy|\rvx)$. This reduces the computation to the estimation of only one entropy and a cross-entropy term:
\begin{align}
I_{r_\rvx}(\rvx;\rvy) &= \E[\log r(\rvy|\rvx)] + H(\rvy).
\label{eq:I_r}
\end{align}
Popular modeling choices for joint and conditional proposals include transforming simple densities
 using normalizing flows \citep{rezende2015variational, kingma2016improving, dinh2017density} or using variational lower-bounds \citep{kingma2013autoencoding}.
 Despite recent advances \citep{durkan2019neural, ho2020denoising}, generative approaches tend to show a trade-off between flexibility and computational costs, which may limit their effectiveness in high-dimensional settings.

\subsection{A Hybrid Method}
We can combine the advantages of both methods by (i) considering proposals $r(\rvx, \rvy)$ such that $\KL(p(\rvx,\rvy)|| r(\rvx,\rvy)) < \KL(p(\rvx,\rvy)||p(\rvx)p(\rvy))$ to lower the variance for the estimates of $\lunnorm{f}{p(\rvx,\rvy)}{r(\rvx,\rvy)}$.
Furthermore, we choose a flexible $f(\rvx, \rvy)$, e.g., a neural network, that refines the proposal.

Both terms of \Cref{eq:general_bound} are now nonzero.
Note that if we already had access to flexible density estimators, we can now simply include them as $r(\rvx, \rvy)$, where $f(\rvx, \rvy)$ corrects for any lack of flexibility.
One can therefore apply an iterative method:
\begin{enumerate}
    \item Pick the best proposal $\hat r(\rvx,\rvy)\in \gR$ to minimize the variational gap. This is equivalent to maximizing $I_r(\rvx,\rvy)$:
    \begin{align}
        \hat r(\rvx,\rvy) \defined \argmax_{r\in\gR} I_r(\rvx;\rvy) 
    \end{align}
    \item Learn the ratio between the best proposal and the joint distribution:
    \begin{align}
        \hat f(\rvx,\rvy) = \argmax_{f\in\gF} \lunnorm{f}{p(\rvx,\rvy)}{\hat r(\rvx,\rvy)}
        \label{eq:energy_objective}
    \end{align}
\end{enumerate}
In practice, these two terms can be jointly optimized for fixed distributions of $\rvx$ and $\rvy$ as long as the objective in \Cref{eq:energy_objective} is treated as constant w.r.t the proposal.

Since our hybrid approach includes a flexible $f(\rvx, \rvy)$, we design in the next section a simple family of proposals $\gR_{Q}$ that are guaranteed to lie closer to the joint distribution $p(\rvx, \rvy)$ than the product of marginals, but yield simple and efficient computation of $I_{r_Q}(\rvx,\rvy)$.

\section{Predictive Quantization}
\label{sec:predictive_quantization}
\begin{algorithm}[t]
   \setstcolor{ggreen}
\begin{algorithmic}
   \STATE {\bfseries input}  $p(\rvx,\rvy)$, \textcolor{ggreen}{$Q(\rvx)$, $s_\psi(Q(\rvx)|\rvy)$}
   \REPEAT
   \STATE \textcolor{ggreen}{$\bar\vx \sim p(Q(\rvx))$}
   \STATE $(\vx_i, \vy_i)_{i=1}^{B} \stackrel{B}{\sim}\ $\st{$ 
p(\rvx,\rvy)$}$\ p(\rvx,\rvy|Q(\rvx)=\bar\vx)$.
    \STATE $\hat{I}(\rvx;\rvy) \gets$ DiscriminativeMIEstimate$((\vx_i, \vy_i)_{i=1}^B; \phi)$
   \STATE \textcolor{ggreen}{$\hspace{7em}  +H(Q(\rvx)) + \log s_\psi(\bar x|(\vy_i)_{i=1}^B)$}
   \STATE $\phi \gets \eta \frac{\partial \hat{I}(\rvx;\rvy)}{\partial \phi}$
   \STATE \textcolor{ggreen}{$\psi \gets \eta \frac{\partial \hat{I}(\rvx;\rvy)}{\partial \psi}$} 
   \UNTIL{converged}
\end{algorithmic}
\caption{Hybrid Mutual Information Estimate with PQ}
\label{alg:pq}
\end{algorithm}

We are interested in proposal distributions that we can easily sample (to estimate $Z_f$) and whose $I_{r}(\rvx, \rvy)$ we can effectively approximate.
We define a joint distribution $r_Q(\rvx, \rvy)$ that factorizes as 
\begin{align}
    r_Q(\rvx,\rvy)\defined p(\rvx)p(\rvy|Q(\rvx)),
\end{align}
where $Q$ maps to some discrete quantization $\bar\rvx=Q(\rvx)$, e.g., a cluster index.
This proposal allows tractable estimation of $I_r(\rvx;\rvy)$ since
\begin{align}
    I_{r_Q}(\rvx;\rvy) &= \E\left[\log\frac{ p(\vy|\bar{\vx})}{p(\vy)}\right]=I(\bar{\vx};\rvy)
    \label{eq:quantized_info}\\
    &= \E\left[\log p(\bar{\vx}|\vy)\right]+ H(\bar{\vx})\nonumber\\
    &\ge \E\left[\log s_\psi(\bar{\vx}|\vy)\right]+ H(\bar{\vx}).\nonumber
\end{align}
Here, $s_\phi(\rvx | \rvy)$ is a variational approximation to $p(\bar{\rvx}|\rvy)$.
This form of the proposal has three key advantages:
\begin{enumerate}
    \item Since $p(\bar{\rvx})$ is discrete, we can obtain good estimates of $H(\bar\vx)$ for reasonable numbers of quantization intervals. Further, we easily sample $r_Q(\rvx,\rvy)$.
    \item $s_\psi(\bar{\rvx}|\rvy)$ can be parametrized as a categorical distribution, obtaining a simple prediction task of $\bar{\rvx}$ given $\rvy$. We can amortize the inference using a deep neural network.
    \item $\KL(p(\rvx,\rvy)||r_Q(\rvx,\rvy)) < \KL(p(\rvx, \rvy)||p(\rvx)p(\rvy))$ for any $Q$ such that $I(\rvy; Q(\rvx)) > 0$. 
\end{enumerate}
The last advantage is quickly identified by observing
\begin{align}
    \KL(p(\rvx,\rvy)||r_Q(\rvx,\rvy)) &\stackrel{Eq. \ref{eq:mi_lbound}}{=} I(\rvx;\rvy)-I_{r_Q}(\rvy;\rvx) \nonumber \\
    &\stackrel{Eq.\ref{eq:quantized_info}}{=} I(\rvx;\rvy)-I(\rvy;Q(\rvx)) \nonumber \\
    &\le I(\rvx;\rvy).
\end{align}
Generally, we want to choose $Q(\rvx)$ so that $I(\rvy; \bar{\rvx})$ is large, causing $r_Q(\rvx, \rvy)$ to be closer to the optimal proposal (visualized in \Cref{fig:attribute_effect}).
On the other hand, the number of quantization intervals should be much smaller than the number of samples
we have to estimate the information.
This is because estimates of discrete entropy $H(\bar\vx)$ become less reliable for a large number of quantization intervals.  
Examples of viable quantization functions could include clustering algorithms
or handcrafted functions of $\rvx$, e.g., a weak label from weakly supervised tasks \citep{ratner2017snorkel}.

The PQ generative estimator can be easily integrated with existing discriminative mutual information estimators by sampling batches $(\vx_i,\vy_i)_{i=1}^B$ in which each $\vx_i$ lies in the same quantized region ($\, \forall i,j\in[B],\, Q(\vx_i)=Q(\vx_j)$) and adding the estimation of $I(Q(\rvx);\rvy)$ to the original discriminative estimate for $KL(p(\rvx,\rvy)||r_Q(\rvx,\rvy))$. In \Cref{alg:pq} we underline (in green) the simple modifications required to integrate PQ with a given discriminative estimator that produces samples from the marginal $p(\rvx)p(\rvy)$ by shuffling the $(\vx_i, \vy_i)$ pair within a batch. By sampling each batch conditioned on one quantized value $\bar\vx$, the shuffling operation becomes equivalent to sampling from $p(\rvx)p(\rvy|\bar\vx)$ instead.

\begin{table*}
    \centering
    \small{
        \begin{tabular}{l|c|c|c}
             &  $r(\rvx,\rvy)$ & $I_r(\rvx;\rvy)$ & Requirements\\\hline
             BA \citep{barber2003im} & $p(\rvx)r_\theta(\vy|\vx)$ & $\E[\log r_\theta(\vy|\vx)]+H(\rvy)$ & $H(\rvy)$ known\\
             DoE \citep{mcallester2020formal} & $p(\rvx)r_\theta(\rvy|\rvx)$ & $\E[\log r_\theta(\vy|\vx)-\log s_\xi(\vy)]$ & $p(\rvy)$ fixed \\
             GM \citep{song2020understanding} & $r_\theta(\vx,\vy)$ & $\E[\log r_\theta(\vx,\vy)-\log s_\psi(\vx)-\log s_\xi(\vy)]$& $p(\rvx)$, $p(\rvy)$ fixed\\
             PQ (\textbf{Ours}) & $p(\rvx) p(\rvy|\bar\rvx))$ & $\E[\log s_\psi(\bar\vx|\vy)] + H(\bar\rvx))$ & $\bar\rvx=Q(\rvx)$
        \end{tabular}
    }
    \caption{Overview of the main generative approaches to mutual information estimation and their requirements. }
    \label{tab:generative}
\end{table*}

\section{Related Work}

Recent work on variational mutual information estimation
\cite{poole2019variational} provides an overview of tractable and scalable objectives for estimating and optimizing Mutual Information (MI), identifying some of the characteristics and limitations of the two approaches. 
The authors analyze the trade-off between bias and variance for discriminative MI estimators focusing on critic architectures and the techniques used to estimate the normalization constant. They propose an interpolation between discriminative estimators with a low variance but high bias \citep{oord2018representation}, and high-variance low-bias estimators \citep{nguyen2010estimating}.

\citet{song2020understanding} show that discriminative approaches such as Mutual Information Neural Estimation (MINE) \citep{belghazi2018mutual} and the Nguyen-Wainright-Jordan (NWJ) method \citep{nguyen2010estimating} exhibit variance that grows exponentially with the true amount of underlying information, making them less suitable for estimation of large information quantities. 
The authors propose to control the bias-variance trade-off of the aforementioned estimators by clipping the critic's values during training.
Other closely related work focuses on the statistical limitation of MI estimators \cite{mcallester2020formal}. The authors underline that the exponential variance scaling affects any variational lower bound and focus on generative mutual information estimation based on the difference of entropies (DoE).

Other works consider factorizing the discriminative approaches using the sum of conditional and marginal MI terms \citep{sordoni2021decomposed}, expressing the density ratio as a sum of several simpler critics \citep{rhodes2020telescoping}, and using Fenchel-Legendre transform for MI estimation and optimization \citep{guo2022tight} to reduce the dependency on large-batch training and increase bound tightness.

Our work further unifies these families of estimators, addressing the exponential scaling issue of discriminative estimators by bringing the distribution used to estimate the normalization constant closer to the joint distribution of the two variables.
The effectiveness of the combination of normalized (generative) and unnormalized (discriminative) distributions has been shown in the context of energy-based models \citet{gao2020flow, xiao2021vaebm}.
We further propose Predictive Quantization (PQ) as a simple yet effective generative estimator inspired by the literature on discrete mutual information estimation \citep{cover2006elements, gao2017estimating} that can be easily combined with any of the aforementioned discriminative estimators to effectively reduce their variance.

\section{Empirical Evaluation}

We test our proposed hybrid models on a particularly challenging correlated mixture of normal distributions, and a discrete-time particle simulation task. 
These tasks have been selected to satisfy the following criteria.
\begin{enumerate}
    \item Known True Mutual Information. Some of the estimators considered in this analysis are lower bounds, while others tend to overestimate the true value of mutual information. For this reason, we exclusively selected benchmarks in which the true value of mutual information is known.  
    \item Controllable Mutual Information. We consider tasks where the amount of information can be tuned by, e.g., a higher dimensionality or increasing the number of particles. 
    We designed our experiments to make the difference between benchmarked estimators explicit by varying the batch size and target mutual information.
\end{enumerate}
    \begin{figure*}
        \centering
        \begin{minipage}{.38\textwidth}
            \centering
            $(a)$
            \includegraphics[width=\textwidth]{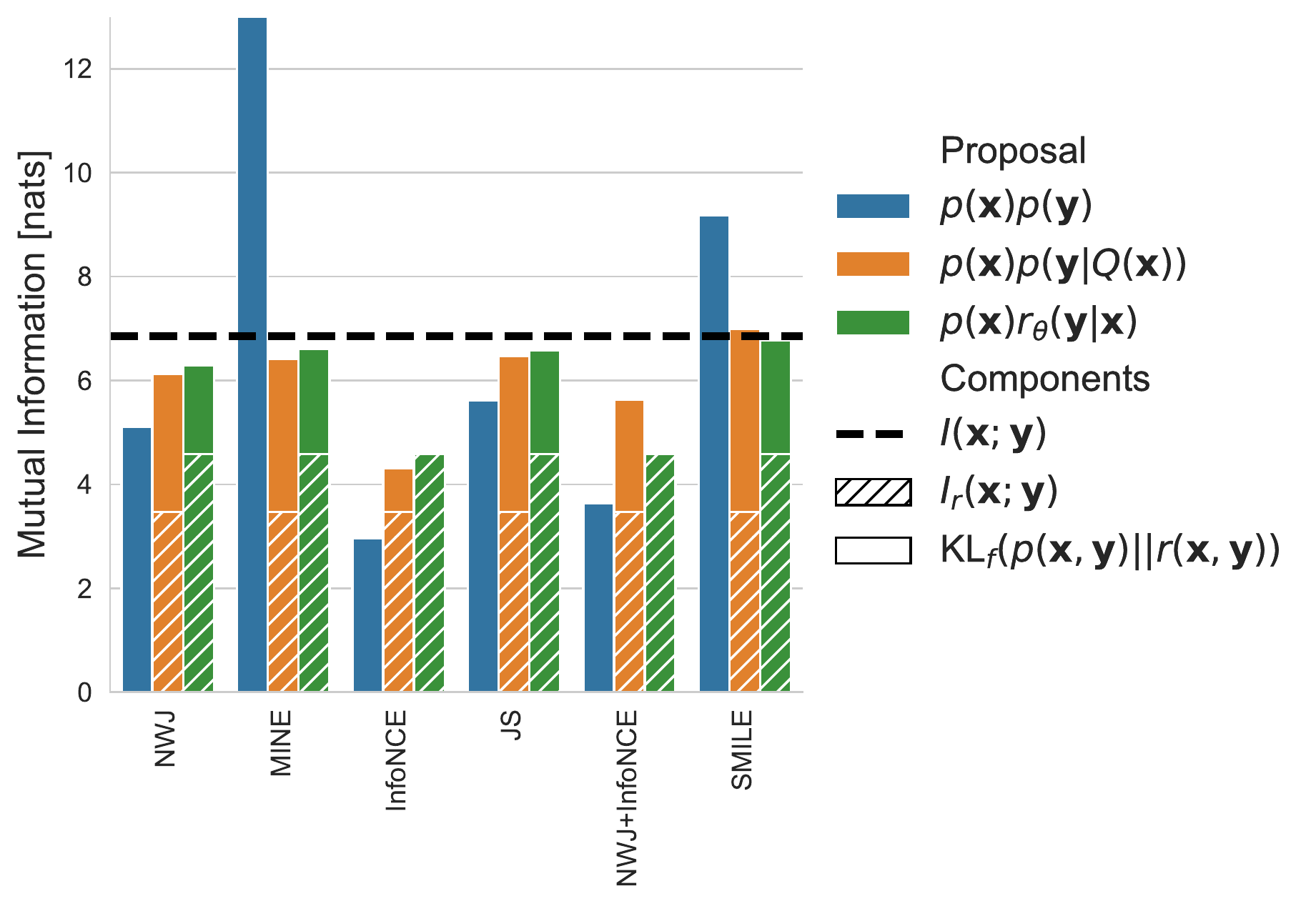}
        \end{minipage}
        \begin{minipage}{.61
\textwidth}
\centering
            $(b)$
            \includegraphics[width=\textwidth]{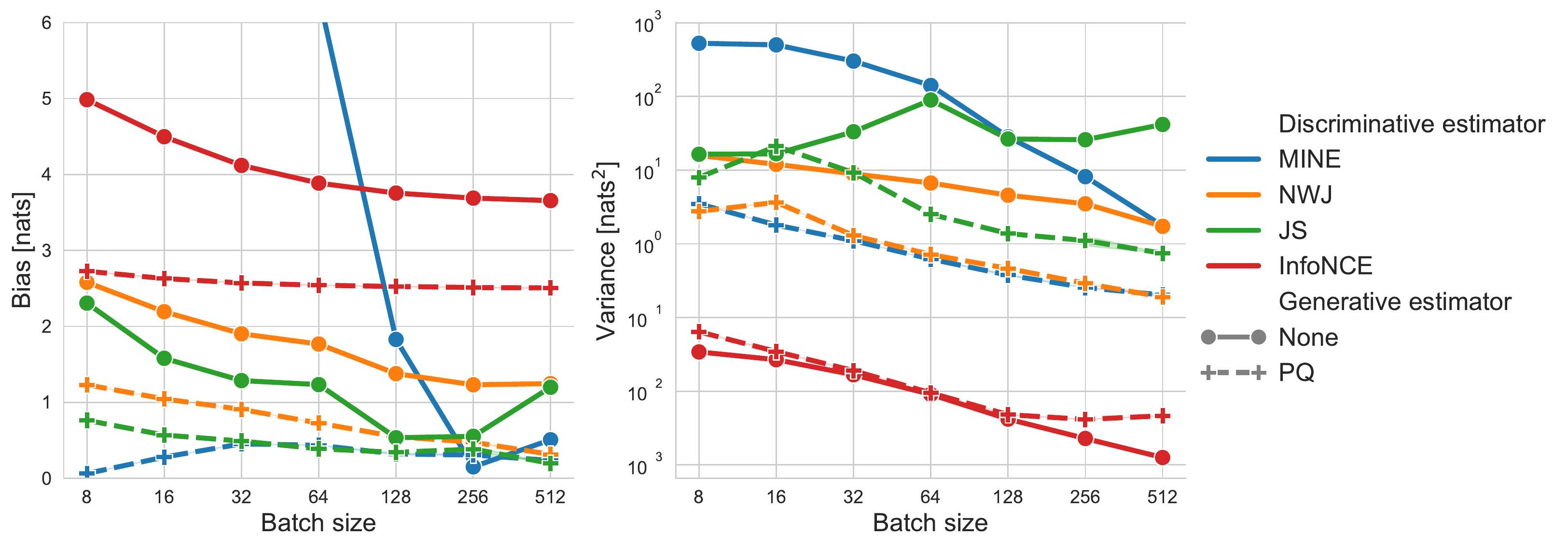}
        \end{minipage}
        \caption{(a) Blue: the estimated mutual information in nats of several discriminative estimators. Orange: mutual information estimates using our PQ method. Green: results obtained by including a learned conditional normal proposal. Hybrid estimators improve upon their discriminative counterparts. We further indicate the parts of the total information contributed by $I_r(\rvx;\rvy)$ and $\lunnorm{f}{p(\rvx,\rvy)}{r(\rvx,\rvy)}$, respectively.
 (b) Bias and variance of the included discriminative estimators (denoted by colors) as a function of batch size. For all estimators, we report a lowered bias using PQ. 
 }
        \label{fig:normal_mixture_batch}
    \end{figure*}
    
\subsection{Models and Optimization}
We evaluate the performance of several discriminative mutual information estimators that differ for the computation of the normalization constant, the parameterization of the energy function, and their optimization (see \Cref{app:discriminative_estimators} for details regarding each estimator).
For each discriminative model, we consider three proposals:
    \begin{enumerate}
        \item The product of the marginals $p(\rvx)p(\rvy)$. This corresponds to using no generative component.
        \item The product of the marginal $p(\rvx)$ and a conditional Normal proposal $\mathcal{N}(\rvy|\boldsymbol\mu_\theta(\rvx),\boldsymbol\sigma^2_\theta(\rvx))$ (BA, DoE). The functions $\boldsymbol\mu_\theta(\rvx)$ and $\boldsymbol\sigma^2_\theta(\rvx)$ are parametrized by neural networks. Note that mutual information estimation using this approach requires estimating the entropy of the marginal distribution $p(\rvy)$.
        \item The product of $p(\rvx)$ and $p(\rvy|Q(\rvx))$ defined in section~\ref{sec:predictive_quantization}. This requires specifying a fixed quantization function $Q(\rvx)$ and a classifier $s_\psi(Q(\rvx)|\rvy)$.
    \end{enumerate}

For a fair comparison, all the architectures used in this analysis consist of simple Multi-Layer Perceptrons (MLP), ReLU activations \cite{nair2010rectified}, and the same neural architecture. All models are trained using Adam \citep{kingma2014adam} with a learning rate of 5e-4 for a total of 100,000 iterations.
    \begin{figure*}
        \centering
        \begin{minipage}{.38\textwidth}
            \centering
            $(a)$
            \includegraphics[width=\textwidth]{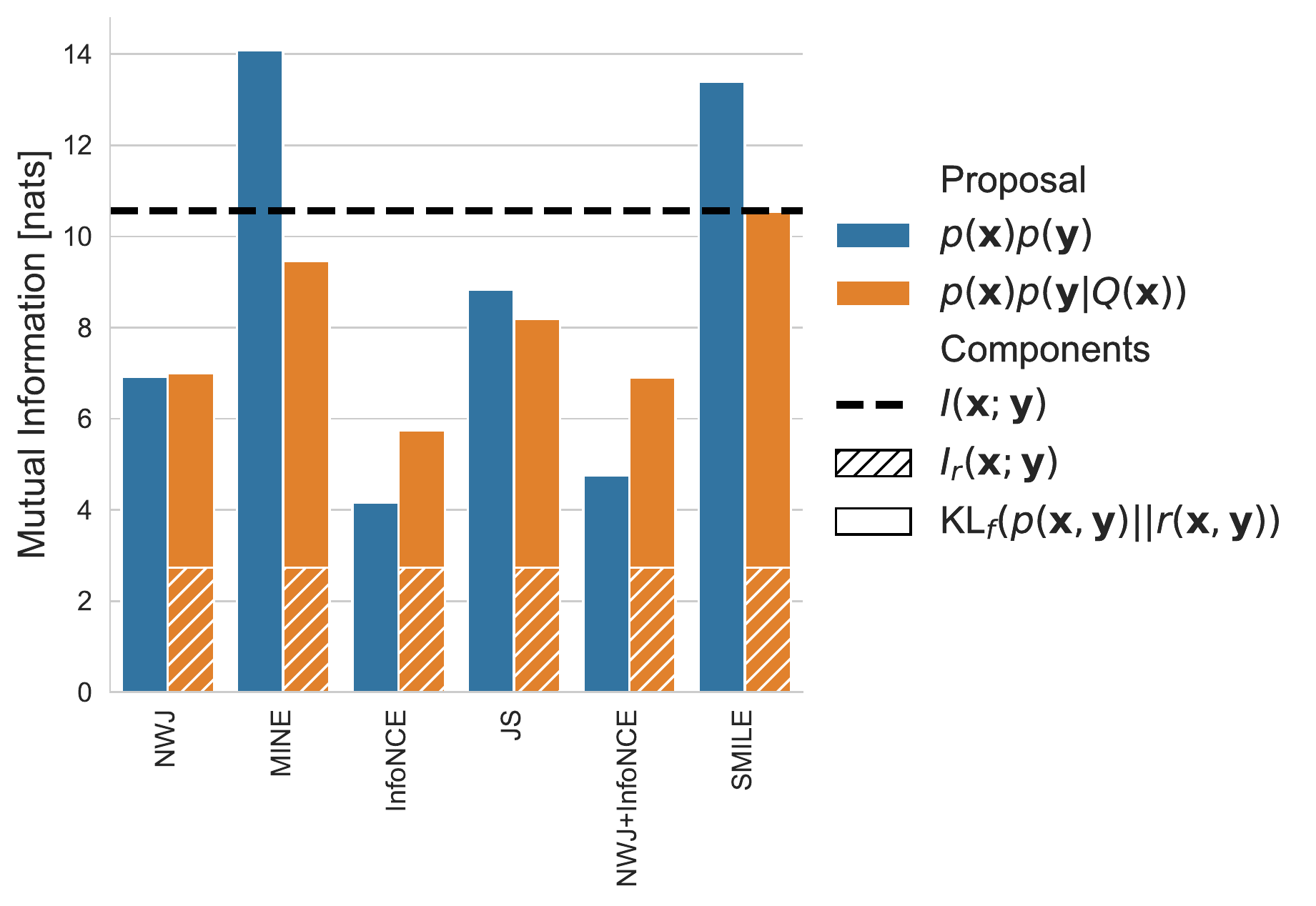}
        \end{minipage}
        \begin{minipage}{.61
\textwidth}
\centering
            $(b)$
            \includegraphics[width=\textwidth]{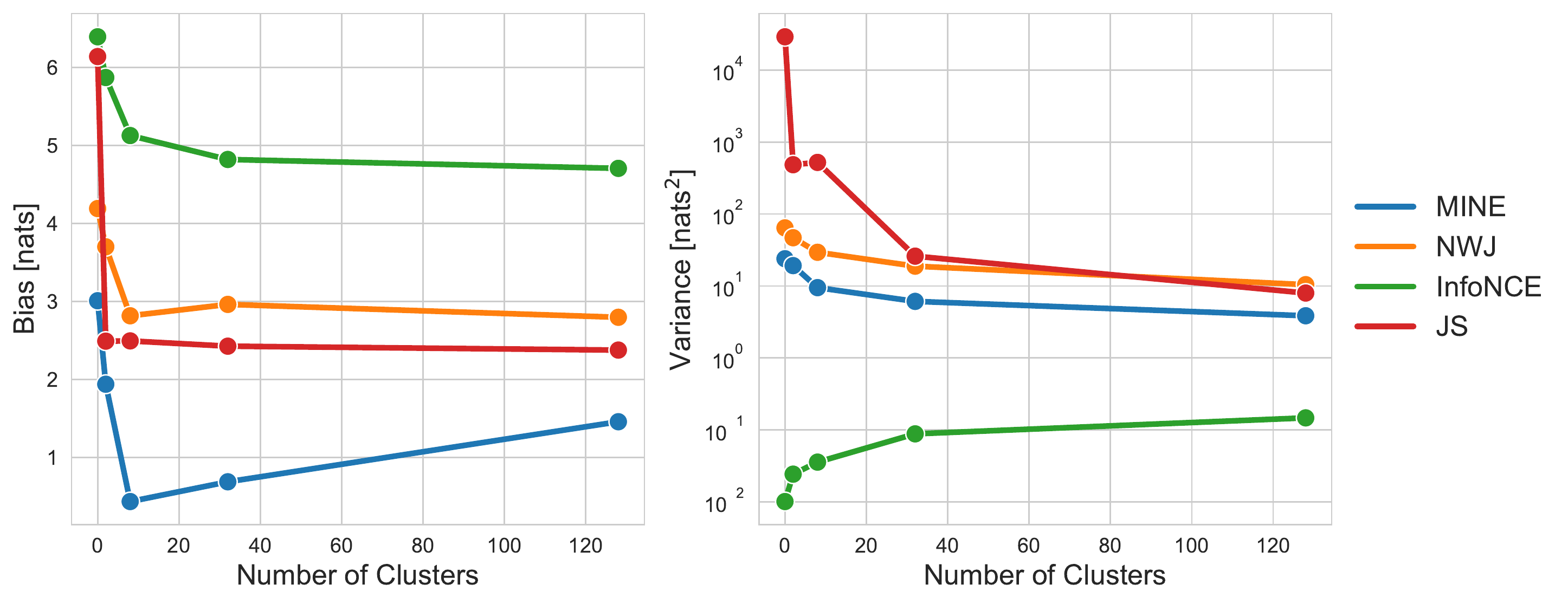}
        \end{minipage}
        \caption{(a) Shows the mutual information estimates of several approaches on the particle dataset. Using PQ (orange), all estimators yield better estimates closer to the true information than the baselines (blue). (b) Bias-variance analysis of the number of clusters used for PQ. Note that 0 clusters correspond to no generative estimator.}
        \label{fig:particles_clusters}
    \end{figure*}

\subsection{Tasks and Results}
    \paragraph{Mixture of Correlated Normals}

    Following previous work \citep{poole2019variational, song2019understanding, guo2022tight}, we create a dataset by sampling $100\,000$ points from a correlated distribution. 
    Instead of using a two-dimensional correlated normal distribution, which simple generative estimators can easily fit, we use a mixture of 4 correlated normal distributions. 
    The location and scale of each component are chosen such that both marginals $p(\rvx)$, $p(\rvy)$ and conditionals $p(\rvy|\rvx)$, $p(\rvx|\rvy)$ are bimodal distributions. 
    The log-density of the product and joint distribution is visualized in \Cref{fig:attribute_effect}.
    Each pair of dimensions shares $\approx1.37$ nats of information. 
    We stacked 5 independent versions to reach an amount of mutual information that is challenging enough to underline issues of the models in this analysis.
    We define the quantization as a component-wise indicator function $Q_2(\rvx)=\mathbb{I}(\rvx>0)$, separating positive and negative values of $\rvx$ for each dimension. 
    
    \Cref{fig:normal_mixture_batch}(a) shows the amount of information estimated by several combinations of generative and discriminative estimators for a fixed batch size of 64 at the end of the training procedure.
    We note that for all analyzed models, adding a generative component results in improved (i.e., less biased) mutual information estimates compared to the purely discriminative baselines (in blue). 
    This is observed for both estimators that underestimate and overestimate the information and holds for both normal proposals (in green) and proposals that use $r(\rvy|Q_2(\rvx))$ (orange).
    However, it is worth mentioning that $r(\rvy|Q_2(\rvx))$ does not require access to the entropy $H(\rvy)$.
    We further indicate the parts of the total information contributed by $I_r(\rvx;\rvy)$ and $\lunnorm{f}{p(\rvx,\rvy)}{r(\rvx,\rvy)}$, respectively.

    \Cref{fig:normal_mixture_batch}(b) reports the effect of increasing the batch size on the bias and variance of the mutual information estimators.
    The addition of a generative component has a similar effect to an increased batch size. 
    In particular, for estimations that suffer from high variance (such as MINE and NWJ) the adding the PQ generative component is comparable to increasing the batch size by a factor of 64. Estimators based on InfoNCE that are characterized by lower variance can benefit from the addition of a generative component to lower their bias. 
    Further analysis of the effect of the number of samples from the proposal are reported in \Cref{app:extra_normal_res}.
    
    \paragraph{Discrete Time Multi-Particle Simulation} 
    Stochastic processes characterize relevant problems in scientific discovery.
    For example, it can be of interest in a dynamical system (e.g., fluid dynamics or molecular dynamics) estimating how much information gets propagated through time.
    We simulate a dataset that aims to resemble the characteristics of a system of particles moving in a fixed energy landscape following discrete-time overdamped Langevin dynamics \cite{ermak1974equilibrium}: $
        \rvx_{t} = \rvx_{t-1} - \epsilon\nabla U(\rvx_{t-1}) + \sqrt{2\epsilon/\beta}\boldsymbol{\eta},
    $
    where $\nabla U(\rvx_t)$ refers to the gradient of an energy $U(\rvx_t)$ evaluated at a position $\rvx_t$, and $\boldsymbol\eta$ is time-independent noise sampled from a unit Normal $\mathcal{N}(\boldsymbol{0},\mathbf{1})$.
    When the system is in equilibrium, $\rvx_t$ follows the Boltzmann distribution $p(\rvx_t)\propto e^{-\beta U(\rvx_t)}$. 
    Therefore, it is possible to compute a good approximation of the entropy. 
    Similarly, the conditional entropy of the transition distribution can be computed as a function of $\epsilon$ and $\beta$. 
    As a result, in this simplified system, we can compute a faithful approximation of mutual information between the positions of a particle at consecutive time steps. We simulate a collection of five independent particles on a two-dimensional energy landscape characterized by multiple wells to create a ten-dimensional trajectory vector $\rvx_t$ for which $I(\rvx_t;\rvx_{t+1})\approx 10.561$ nats.
    Furthermore, to introduce spurious correlation, we first concatenate 10-dimensional time-independent Normal noise and 10 constant dimensions before applying an invertible non-linear function. This last operation increases the dimensionality to 30 without effectively changing the value of mutual information. Further details on the dataset creation procedure can be found in \Cref{app:extra_particles}.

    To produce the quantization function used for this experiment, we first project the 30-dimensional feature space onto 10 principal components using Temporal Independent Component Analysis (TICA) \citep{molgedey1994separation}. 
    TICA ensures that particles that correlate highly through time, have similar representations.
    Secondly, we apply a K-nearest neighbors clustering to obtain $Q(\rvx)$ for all particles and time-steps of the dataset. 
    The rationale behind this choice of the quantization function is to make sure that $Q(\rvx)$ captures temporal information by grouping those particles. 
    Thereby, it yields a proposal distribution $r_Q(\rvx_{t+1}, \rvx_t)$ that improves $I(\rvx_{t}; Q(\rvx_{t+1}))$.

    \Cref{fig:particles_clusters}(a) shows the mutual information result for discriminative (blue) and hybrid models using PQ as the generative components (orange). Overall, we observe that including a generative component yields improved mutual information estimates of the tested estimators. 
    \Cref{fig:particles_clusters}(b) shows the effect of increasing the number of quantization intervals $Q(\rvx)$ on the bias-variance trade-off.
    In this case, this number is provided by the clusters for the TICA quantization. 
    Increasing the number of clusters results in a variance reduction for most of the estimators, except for InfoNCE-based estimators. 
    This is because InfoNCE estimators are characterized by low variance and high bias, and the increase in variance is due to the estimation of the PQ generative component. 
    We also notice that the bias for some estimators tends to increase for large numbers of quantized values. 
    We hypothesize that this effect is due to the induced scarcity of the samples from $p(\rvy|Q(\rvx))$.
    
\section{Conclusion}
We introduced a hybrid approach for mutual information estimates that generalizes discriminative and generative methods and combines advantages of both approaches.
On top of that, we propose Predictive Quantization (PQ): a simple generative method that can be used to improve discriminative estimators.
These contributions analytically yield improved mutual information estimates with lower variances.
Theoretical results were confirmed experimentally on two challenging mutual information estimation tasks.

\paragraph{Limitations and Future Work}
Although the hybrid approaches proposed in this work help to address limitations of generative and discriminative approaches in the literature, they introduce additional complexity due to the interplay between the two components. 
Nevertheless, we believe that simple (or non-parametric) proposals could be used together with discriminative models to maximize information between learned representations in future work.

\newpage
\nocite{langley00}

\bibliography{bib}
\bibliographystyle{icml2023}

\newpage
\appendix
\onecolumn
\section{Notation and conventions}

\label{app:notation}
\begin{enumerate}
    \item Random variables
    
    We use $\rvx$ to denote random variables, while $\vx$ refers to their realization. We use $p(\vx)$ as a shorthand for $p(\rvx=\vx)$ and $p(\rvx|\vy)$ for $p(\rvx|\rvy=\vy)$.
    
    \item Expectations
    
    We use $\E_{\vx\sim p(\rvx)}[f(\vx)]$ to denote the expected value of $f(\vx)$ with respect to the density $p(\rvx)$ for both discrete and continuous random variables. We omit the subscript $\E[f(\vx)]\defined\E_{\vx\sim p(\rvx)}[f(\vx)]$ for brevity. When no subscript is specified, the expectation is considered with respect to the joint density $p$.
    
    \item Kullback-Leibler divergence
    We use $\KL(p(\rvx)||q(\rvx))\defined\E\left[\log\frac{p(\vx)}{q(\vx)}\right]$ to denote the Kullback-Leibler (KL) divergence between the densities $p(\rvx)$ and $q(\rvx)$. 
    
    \item Mutual information and entropy
    
    We use the notation $I(\rvx;\rvy)$ to indicate the value of mutual information between the random variables $\rvx_1$ and $\rvy$ (see definition in equation~\ref{eq:mi_def}). 
    
    We use $H(\rvx)\defined \E[-\log p(\vx)]$ to denote the entropy of $\rvx$ for both discrete and continuous (differential entropy) random variables.
\end{enumerate}

\section{Discriminative Mutual Information estimators}
\label{app:discriminative_estimators}
We report a summary of the objectives and parametrized critic architectures $g_\phi(\vx,\vy)$ for a batch $(\vx_i,\vy_i)_{i=1}^B$ of samples from $p(\rvx,\rvy)$ and with $(\vy_{i,j}')_{j=1}^K$ representing $K$ samples from the proposal $r(\rvy|\vx_i)$ for each $\vx_i$. 
\begin{itemize}
    \item NWJ \citep{nguyen2010estimating}:
    \begin{align}
            \frac{1}{B} \sum_{i=1}^B g_\phi(\vx_i,\vy_i)-\frac{1}{e\ BK} \sum_{i=1}^B\sum_{j=1}^K e^{g_\phi(\vx_i,\vy_{i,j}')+1}
    \end{align}
    \item MINE \citep{belghazi2018mutual}
        \begin{align}
            \frac{1}{B} \sum_{i=1}^B g_\phi(\vx_i,\vy_i)-\log\frac{1}{BK} \sum_{i=1}^B\sum_{j=1}^K e^{g_\phi(\vx_i,\vy_{i,j}')}
        \end{align}
        A running average of $\frac{1}{BK} \sum_{i=1}^B\sum_{j=1}^K e^{g_\phi(\vx_i,\vy_{i,j}')+1}$ is used for training.
    \item InfoNCE \cite{oord2018representation} 
    \begin{align}
        \frac{1}{B} \sum_{i=1}^B g_\phi(\vx_i,\vy_i)-\frac{1}{B}\sum_{i=1}^B\log\frac{1}{K} \sum_{j=1}^K e^{g_\phi(\vx_i)^T\vy_{i,j}'},
    \end{align}
    with $\vy_{i,j}=\vy_j$ and $B=K$.
    \item JS \cite{devon2019learning, poole2019variational}
    Mutual information estimation is performed analogously to NWJ, the critic is optimized by minimizing
    \begin{align}
        \frac{1}{B} \sum_{i=1}^B -s^+(-g_\phi(\vx_i,\vy_i))-\frac{1}{BK} \sum_{i=1}^B\sum_{j=1}^K -s^+(g_\phi(\vx_i,\vy_{i,j}')),
    \end{align}
    in which $s^+$ corresponds to the softplus function.
    \item NWJ+InfoNCE \cite{poole2019variational}
    This estimator uses an interpolation of the baseline used in NWJ and InfoNCE:
    \begin{align}
        \frac{1}{B} \sum_{i=1}^B \left[g_\phi(\vx_i,\vy_i)- \log 
 B(\vx_i)\right] -\frac{1}{BK} \sum_{i=1}^B\sum_{j=1}^K \frac{e^{g_\phi(\vx_i,\vy_{i,j}')}}{B(\vx_i)}
         + 1,
    \end{align}
    with $B_\alpha(\vx_i)\defined 
        \alpha + (1-\alpha)\frac{1}{K} \sum_{j=1}^K e^{g_\phi(\vx_i,\vy_{i,j}')}$.
    Throughout our experiments, we use $\alpha=0.5$
    \item SMILE\cite{song2020understanding}
    Mutual Information is estimated equivalently to MINE with the exception of value clipping $CLIP$ inside the exponential
        \begin{align}
            \frac{1}{B} \sum_{i=1}^B g_\phi(\vx_i,\vy_i)-\log\frac{1}{BK} \sum_{i=1}^B\sum_{j=1}^K e^{CLIP(g_\phi(\vx_i,\vy_{i,j}'),\tau, \tau)}.
        \end{align}
        The parameters $\phi$ are updated using the same objective as JS.

        Throughout the experiments, we used the value $\tau=5$.
\end{itemize}

\section{Proofs}
\label{app:proofs}

\subsection{Composing $I(\rvx; \rvy)$ from $I_r(\rvx;\rvy)$ and $\KL(p(\rvx, \rvy)||r(\rvx, \rvy)$}
\label{app:ir_kl_composition}
\begin{align}
    I_r(\rvx;\rvy) + \KL(p(\rvx, \rvy)||r(\rvx, \rvy)) &= \mathbb{E}\left[ \log \frac{r(\vx, \vy)}{p(\vx)p(\vy)} \right] + \mathbb{E}\left[\log \frac{p(\vx, \vy)}{r(\vx, \vy)}\right] \nonumber \\
    &= \mathbb{E}\left[ \frac{p(\vx, \vy)}{p(\vx)p(\vy)} \right] \nonumber \\
    &= I(\rvx;\rvy)
\end{align}

\subsection{Variance}
We report the proof to show the exponential growth of the variance of the normalization constant for completeness. Similar proofs can be found in \cite{mcallester2020formal, song2020understanding}.
\begin{align}
    \Var_{r}\left[e^{f(\vx,\vy)}\right]& \ge \E_{r}\left[\left(e^{f(\vx,\vy)}\right)^2\right]-\E_{r}\left[e^{f(\vx,\vy)}\right]^2\nonumber\\
    &= Z^2\E_{r}\left[\left(\frac{r(\vx,\vy)e^{f(\vx,\vy)}}{r(\vx,\vy)Z}\right)^2\right]-Z^2\nonumber\\
    &= Z^2\iint r(\vx,\vy) \left[\left(\frac{q(\rvx,\rvy)}{r(\vx,\vy)}\right)^2-1\right]d\vx d\vy\nonumber\\
    &\defined Z^2\CHI(q(\rvx,\rvy)||r(\rvx,\rvy))\nonumber\\
    &\ge Z^2\left(e^{\KL(q(\rvx,\rvy)||r(\rvx,\rvy))}-1\right),
    \label{eq:variance}
\end{align}
A proof for the upper bound shown in the last step can be found in \cite{gibbs2002} (Theorem 5).

\subsection{An Improved Bound}
Consider a set of attainable critics $\mathcal{F}$. When we include at least a constant function $f_k: \sX \times \sY \to \{k\}$, where $k$ is a constant, our lower bound is at least as good as a fully generative approach.
\begin{align}
    \max_{(r, f) \in \mathcal{R} \times \mathcal{F}} I_q(\rvx, \rvy) &= \max_{(r, f) \in \mathcal{R}\times \mathcal{F}} \left[ I_r(\rvx, \rvy) + \KL_f(p(\vx, \vy) || r(\vx, \vy)) \right]\\
    &=  \max_{(r, f) \in \mathcal{R} \times \mathcal{F}} \left[I_r(\rvx;\rvy) + \mathbb{E} [f(\vx, \vy)] - \log \mathbb{E}_r \left[ e^{f(\vx, \vy)}\right] \right] \\
    &\stackrel{{f_k \in \mathcal{F}}}{\geq}  \max_{r \in \mathcal{R}} \left[I_r(\rvx;\rvy) + k - \log \mathbb{E}_r \left[ e^{k}\right] \right] \\
    &=  \max_{r \in \mathcal{R}} \left[I_r(\rvx;\rvy)\right] 
\end{align}

\section{Experimental Details}
We include additional details on the models and datasets for reproducibility purposes.
A full version of the code and dataset will be published upon acceptance.

\subsection{Architectures}
Generative and discriminative architectures used for the experiments on the Normal Mixture dataset consist of MLPs with hidden layers of 256 and 128 hidden units with ReLU activations.
The same architecture is used for joint and separable critic architectures, and for parametrizing learnable conditional distribution $r_\theta(\rvy|\rvx)$.
The parametrization for the variational distribution $s_\phi(Q(\rvx)|\rvy)$ used in the PQ model uses a simpler architecture consisting of one hidden layer of 128 units.

\subsection{Normal Mixture}
\label{app:extra_normal_res}
Each dimension of the samples of $\rvx$ and $\rvy$ are drawn from a mixture of four correlated Normal distributions with mixing proportions $\pi = [1/4,1/4,1/4,1/4]$, covariance $\Sigma=\begin{bmatrix} 1 & 0.95\\ 0.95 & 1 \end{bmatrix}$ and mean $\mu_1=[\epsilon + \delta, -\epsilon + \delta]$, $\mu_2=[-\epsilon - \delta, \epsilon - \delta]$, $\mu_3=[\epsilon - \delta, -\epsilon - \delta]$ and $\mu_4=[-\epsilon + \delta, \epsilon + \delta]$.
 The values of bias and variance reported in the figures are computed on the last epoch of the end of the training procedure.

\subsection{Discrete-time Multi-particle Simulation}
\label{app:extra_particles}

\begin{figure}[!ht]
    \centering
    \includegraphics[width=0.5\textwidth]{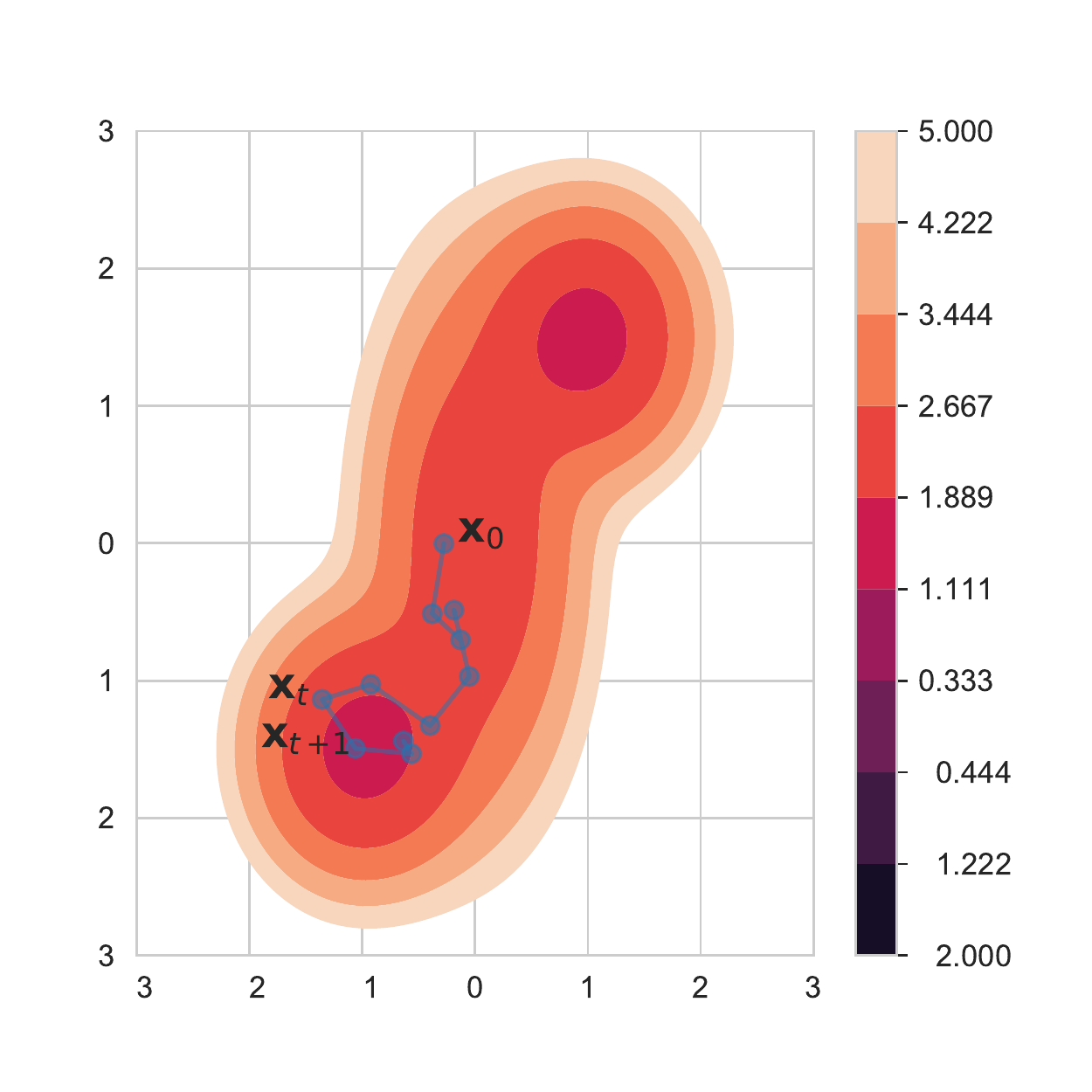}
    \caption{Visualization of the 2D energy $U(\rvx)$ and a portion of simulated trajectory for one particle used to generate the dataset.}
    \label{fig:particle_energy}
\end{figure}

The trajectories for the discrete-time multi-particle simulation task are generated using the log-density of a mixture of Normal as an energy function (\Cref{fig:particle_energy}).
For simulation, we use $\epsilon=0.05$ and $\beta=0.3$. Since the equilibrium distribution generated using discrete timesteps slightly differs from the Boltzmann distribution, we re-estimate the entropy by binning 2M samples into a 100$\times$100 grid. 
Each trajectory is obtained by considering 100.000 consecutive samples after an initial burn-in of 100.000 samples starting from a fixed initialization to ensure convergence to the equilibrium distribution.
After concatenating the trajectories for the 5 particles, we concatenate 10 independent samples from a unit Normal distribution and 10 constant zero dimensions to each timestep. Finally, we apply to layers of randomly initialized affine autoregressive transformations, obtaining a correlated 30-dimensional vector for each timestep.  The values of bias and variance reported in the figures are computed on the last 10 epochs of the training procedure to mitigate the effect of the high variance.

\section{Additional Experimental Results}

\begin{figure}
    \centering
    \includegraphics[width=\textwidth]{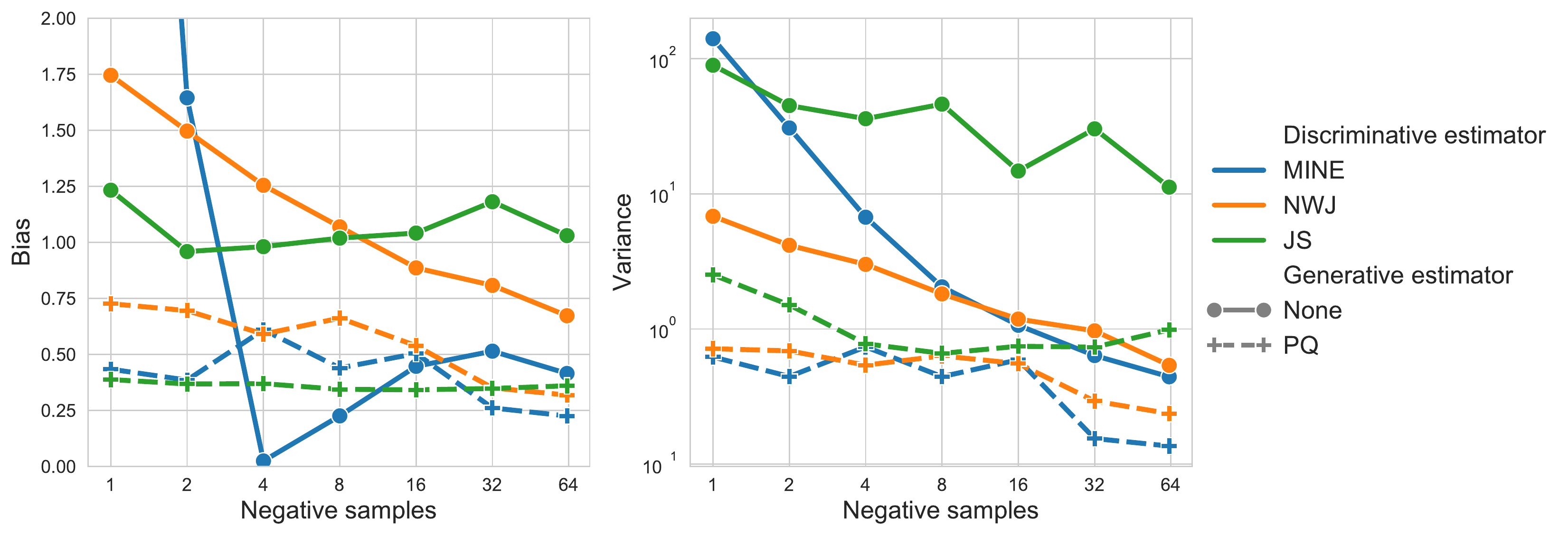}
    \caption{Effect of increasing the number of negative samples on mutual information estimation for a fixed batch-size of 64. Note that the MINE estimator switches from overestimating ($<4$ negatives) to underestimating ($>4$ negatives). }
    \label{fig:normal_negative_samples}
\end{figure}

Figure \Cref{fig:normal_negative_samples} shows the effect of increasing the number of samples from $r(\rvy|\rvx)$ used to estimate the normalization constant for several discriminative mutual information estimators and their hybrid counterparts.
We can observe a similar effect to the one obtained a batch-size (\Cref{fig:normal_mixture_batch}). The introduction of the PQ generative model has a considerable effect on reducing the variance of the estimators: using PQ together with the MINE or NWJ
has a similar effect to increasing the number of negatives by a factor x64.



\end{document}